\documentclass[10pt,letterpaper,twocolumn]{article}

\usepackage{iccv}
\usepackage{times}
\usepackage{epsfig}
\usepackage{graphicx}
\usepackage{amsmath}
\usepackage{amssymb}
\usepackage{pifont}

\usepackage[american]{babel}
\usepackage{url}            %
\usepackage{bm}
\usepackage{microtype}      %
\usepackage{graphicx}
\usepackage[american]{babel}
\usepackage{multirow}
\usepackage{amsmath}
\usepackage{amssymb}
\usepackage{algorithm,algorithmic}
\usepackage{diagbox}
\graphicspath{{figure/}}

\usepackage[margin=4pt,font=small,labelfont=bf,tableposition=top]{caption}

\usepackage[pagebackref=true,breaklinks=true,letterpaper=true,colorlinks,bookmarks=false]{hyperref}

\iccvfinalcopy %

\ificcvfinal\fi
\begin{document}

\title{
  Adversarial PoseNet:
  A Structure-aware Convolutional Network for Human Pose Estimation
}

\author{
  Yu Chen$^{1*}$, Chunhua Shen$^2$, Xiu-Shen Wei$^{3}$\thanks{Y. Chen and X.-S. Wei's 
  contribution was made when visiting The University of Adelaide. Correspondence should be addressed to C. Shen
  (e-mail: chhshen@gmail.com).  
   },
   Lingqiao Liu$^2$, Jian Yang$^1$\\ 
  $^1$Nanjing University of Science and Technology, China\\
  $^2$The University of Adelaide, Australia\\
  $^3$Nanjing University, China
}

\maketitle
\begin{abstract}
For human pose estimation in monocular images, 
joint occlusions and overlapping upon human bodies often result in deviated 
pose predictions. Under these circumstances, biologically implausible pose predictions may be produced. In contrast, human vision is able to predict poses by exploiting geometric constraints of joint inter-connectivity. To address the problem by incorporating priors about the structure of human bodies, we propose a novel structure-aware convolutional network to implicitly take such priors into account during training of the deep network. Explicit learning of such constraints is typically challenging. Instead, we design discriminators to distinguish the real poses from the fake ones (such as biologically implausible ones). If the pose generator (\textit{G}) generates results that the discriminator fails to distinguish from real ones, the network successfully learns the priors.

To better capture the structure dependency of human body joints, the generator \textit{G} is designed in a stacked multi-task manner to predict poses as well as occlusion heatmaps. Then, the pose and occlusion heatmaps are sent to the discriminators to predict the likelihood of the pose being real. Training of the  network follows the strategy of conditional Generative Adversarial Networks (GANs). The effectiveness of the proposed network is evaluated on two widely used human pose estimation benchmark datasets. Our approach significantly outperforms the state-of-the-art methods and almost always generates plausible human pose predictions.
\end{abstract}

\tableofcontents
\clearpage

\begin{figure*}[h!]
\centering
\includegraphics[width=2\columnwidth]{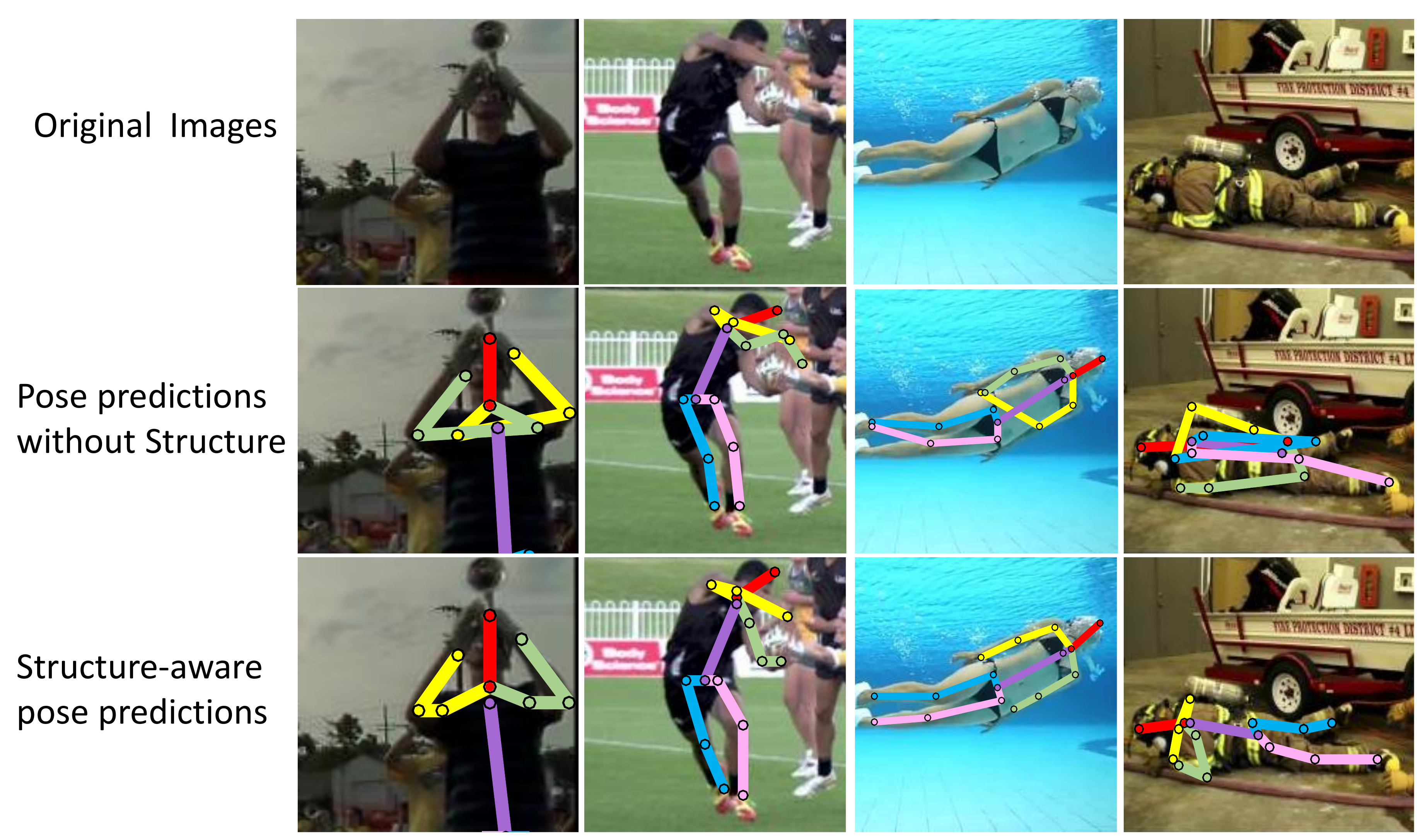}
\caption{\textbf{Motivation}. We show the importance of strongly enforcing priors about the human body structure during  training of DCNNs for pose estimation. Learning without using such priors generates inaccurate results.}
\label{fig:The-blue-shape}
\end{figure*}

\section{Introduction}\label{sec:Introduction}

Human pose estimation is a key step in understanding the actions of people in images and videos. Understanding of a person's limb articulation location is very helpful for high-level vision tasks like human tracking, action recognition, and also serves as a fundamental tool in fields such as human-computer interaction applications. It is a challenging task in computer vision due to high flexibility of body limbs, self and outer occlusion, various camera angles, etc.

Recently, significant improvements have been achieved on this topic by using Deep Convolutional Neural Networks (DCNNs)~\cite{conf/nips/TompsonJLB14,conf/cvpr/TompsonGJLB15,conf/cvpr/ToshevS14,conf/cvpr/ChuOLW16,conf/cvpr/WeiRKS16,conf/eccv/NewellYD16,conf/eccv/BulatT16}. These approaches mainly follow the strategy of regressing heatmaps of each body part using DCNNs. These regression models have shown great ability of learning better feature representations. However, for body parts with heavy occlusions (especially from body parts of surrounding  people) and background which seems similar to body parts, DCNNs may have difficulty in regressing accurate heatmaps.

\begin{figure*}[!t]
\centering
\includegraphics[width=0.985\textwidth]{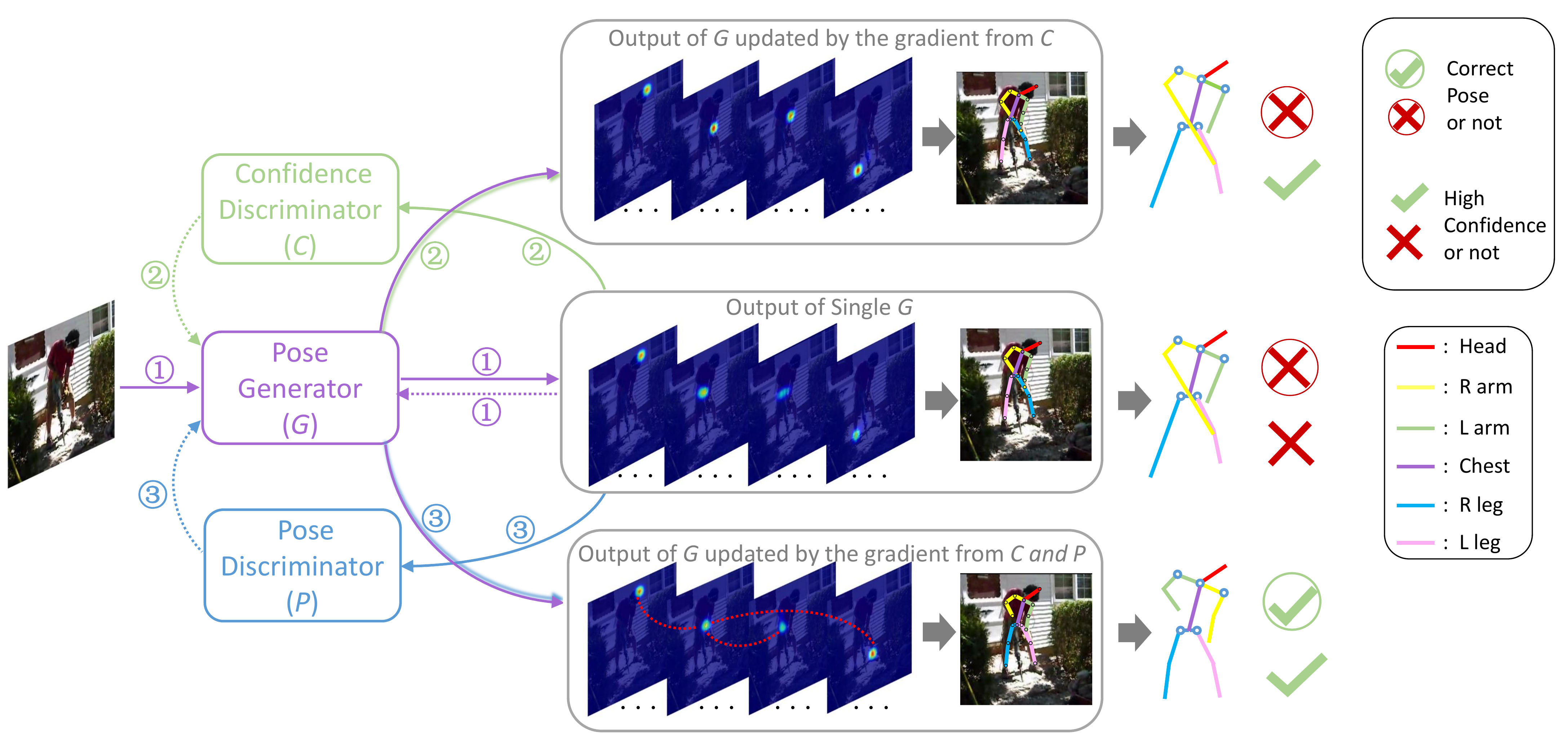}
\caption{Overview of the proposed Structure-aware Convolutional Network for human pose estimation. The sub-network in purple is the stacked multi-task network (\emph{G}) for pose generation. The networks in blue (\emph{P}) and green (\emph{C}) are used to discriminate whether the generated pose is ``real'' (reasonable as a body shape) and whether the generator has strong confidence in locating the body parts, respectively. Dashed lines into \textit{G} indicate backward gradients to update \textit{G}. \textcircled{1} shows the forward and backward of the \textit{G} net. \textcircled{2} shows the process of \textit{G} updated by the gradient from the \textit{C} net. Then, \textit{G} is updated by the gradients from \textit{P} as shown in lines with \textcircled{3}.}
\label{fig:Structure-of-the}
\end{figure*}

Human vision is capable of learning the variety and limitless of human body shape structures from observations. Even under extreme occlusions, we
can infer the potential poses and remove the implausible ones. It is, however, very challenging to incorporate the priors about human body structures into DCNNs, because, as pointed out in~\cite{conf/nips/TompsonJLB14}, the low-level mechanics of DCNNs is typically difficult to interpret, and DCNNs are most capable of learning features.

As a consequence, an unreasonable human pose may be produced by a standard DCNN. As shown in Fig.~\ref{fig:The-blue-shape}, in challenging test cases with heavy occlusions, standard DCNNs tend to perform poorly. To solve this problem, priors about the structure of the body joints must be considered. The key point of this problem is to learn the real body joints distribution from a large amount of training data. However, explicit learning of such a distribution can be very difficult.

To address this problem, we attempt to learn the distribution of the human body structures {\em implicitly}. We suppose that we have a ``discriminator'' which can tell whether the predicted pose is geometrically reasonable. If the DCNN regressor is able to ``deceive'' the ``discriminator'' that its predictions are all reasonable, the network would have successfully learned the priors of the human body structure. Inspired by the recent success in Generative Adversarial Networks (GAN)~\cite{journals/corr/RadfordMC15,journals/corr/ZhaoML16,conf/nips/SalimansGZCRCC16,conf/nips/GoodfellowPMXWOCB14,conf/nips/DentonCSF15}, we propose to design the ``discriminator'' as the discriminator network while the regression network functions as a generative network. Training the generator in the adversarial manner against the discriminator exactly meets our intention.

To accomplish the above goals, the discriminator should be fed with sufficient information to perform classification, while the generator should have the ability in modeling the complicated features in pose estimation. Thus, a multi-task learning network \textit{G} is designed, which simultaneously regresses the
pose heatmaps and the occlusion heatmaps. Based on the pose and occlusion heatmaps, the pose discriminator (\textit{P}) is used to tell whether the body configuration is plausible.

In addition, our preliminary results show that correct locations often correspond to highly confident heatmaps. Therefore, we design another discriminator to
make a decision on the confidence of the predicted pose heatmaps. The generator is asked to ``fool'' both the pose and confidence discriminators by training \textit{G} and $\left\{\emph{P},\emph{C}\right\} $ in the generative adversarial manner. Thus, the human body structure  is implied in the \textit{P} net by  guiding \textit{G} to the direction that is close to ground-truth heatmaps and satisfies joint-connectivity constraints of the human body. The learned \textit{G} net is expected to be more robust to occlusions and cluttered backgrounds where the precise description for different body parts are required.

The main contributions of this work are three folds.
\begin{itemize}
  \itemsep -0.125cm
\item We design a novel network framework for human pose estimation which takes the geometric constraints of human joints connectivity into consideration. By incorporating the priors of the human body, prediction mistakes caused by occlusions and cluttered backgrounds are considerably reduced. Even when the network fails, the outputs of the network appear more like ``human'' predictions instead of ``machine'' predictions.

\item To our best knowledge, we are the first to use Generative Adversarial Networks to exploit  the constrained human-pose distribution for improving  human pose estimation. We also design a stacked multi-task network for predicting both the pose heatmaps and the occlusion heatmaps to achieve better results.

\item We evaluate our method on two public human pose estimation datasets. Our approach significantly outperforms state-of-the-art methods, and is able to consistently produce more plausible pose predictions compared to previous methods.
\end{itemize}

\subsection{Related Work} \label{sec:Related-Work}

Our work is closely related to work using heatmap based DCNN methods for human pose estimation and Generative Adversarial Networks.

\textbf{Human Pose Estimation.} Traditional human pose estimation methods often follow the framework of tree structured graphical model~\cite{journals/ijcv/EichnerMZF12,journals/ijcv/BuehlerEHZ11,conf/eccv/SappTT10,conf/cvpr/YangR11,conf/iccv/PishchulinAGS13,conf/cvpr/SappT13}. With the introduction of ``DeepPose'' by Toshev \emph{et al}.~\cite{conf/cvpr/ToshevS14}, deep network based methods become more popular in this area. This work is more related to the methods generating pose heatmaps from images~\cite{conf/cvpr/YangOLW16,conf/eccv/NewellYD16,conf/cvpr/TompsonGJLB15,conf/cvpr/WeiRKS16,conf/cvpr/ChuOLW16,conf/cvpr/PishchulinITAAG16,conf/eccv/InsafutdinovPAA16,conf/nips/TompsonJLB14}. For example, Tompson \emph{et al}.~\cite{conf/nips/TompsonJLB14} generated heatmaps by running an image through multiple resolution banks in parallel to simultaneously capture features at a variety of scales. Tompson \emph{et al}.~\cite{conf/cvpr/TompsonGJLB15} used multiple branches of convolutional networks to fuse the features from an image pyramid, and used Markov Random Field (MRF) for post-processing. In the following, Convolutional Pose Machine~\cite{conf/cvpr/WeiRKS16} incorporated the inference of the spatial correlations among body parts within convolutional networks. Hourglass Network~\cite{conf/eccv/NewellYD16} 
proposed a state-of-the-art architecture for bottom-up and top-down inference with residual blocks. The structure of our \textit{G} net is also a fully convolutional network with ``conv-deconv'' architecture. However, our network is designed in a multi-task manner with features of both tasks connected to the feature for the second stacked network.

{\bf Generative Adversarial Network.} Generative Adversarial Networks have been widely studied in previous work for discrete labels~\cite{journals/corr/MirzaO14}, text~\cite{reed2016generative} and also images. The image-conditional models have tackled inpainting~\cite{conf/cvpr/PathakKDDE16}, image prediction from a normal map~\cite{conf/eccv/WangG16}, future frame prediction~\cite{journals/corr/MathieuCL15}, future state prediction~\cite{conf/eccv/ZhouB16}, product photo generation~\cite{conf/eccv/YooKPPK16}, and style transfer~\cite{conf/eccv/LifshitzFU16}. Human pose estimation can been considered as a translation from a RGB image to a multi-channel heatmap. The designed bottom-up and top-down \textit{G} net can well accomplish this translation. Different from previous work, the goal of the discrimination network is not only to distinguish the fake from real, but also to incorporate geometric constrain to the model. This is the reason for the different training strategy for fake samples compared with traditional GANs which will be explained in detail in
the following sections.
\section{The Proposed Adversarial PoseNet}\label{sec:GcNet-for-Human}

As mentioned in Fig.~\ref{fig:Structure-of-the}, our Adversarial PoseNet model consists of three parts, \emph{i.e.}, the pose generator network \textit{G}, the pose discriminator network \textit{P} and the confidence discriminator \textit{C}. The generative network is a bottom-up and top-down network, where the inputs are the RGB images and the outputs are 32 heatmaps for each input image in our case. Half of the returned heatmaps are pose estimations for 16 pose key points, and the other half are for the corresponding occlusion predictions. The values in each heatmap are confidence scores in the range of $\left[0,1\right]$ where a Gaussian blur is done around the ground truth position.

Without discriminators, \textit{G} will be updated simply by forward and backward propagations of itself ({cf.}, the lines with \textcircled{1} in Fig.~\ref{fig:Structure-of-the}). That might generate low confidence and even incorrect location pose estimations. It is necessary to leverage the power of discriminators to correct these poor estimations. Therefore, two discriminator networks \emph{C} and \emph{P} are introduced into the framework.

After updating \textit{G} by training with \textit{C} in the adversarial manner (cf. the lines with \textcircled{2}), more confident results are produced. Furthermore, after training \textit{G} with both \textit{P} and \textit{C} (cf. the lines with \textcircled{3}), the human body priors are  implicitly exploited,  and the prediction confidences are accordingly improved.

\subsection{Multi-Task Generative Network}\label{subsec:Multi-task-Generative-Network}

\begin{figure*}[!t]
\centering
\includegraphics[width=1.4600005\columnwidth]{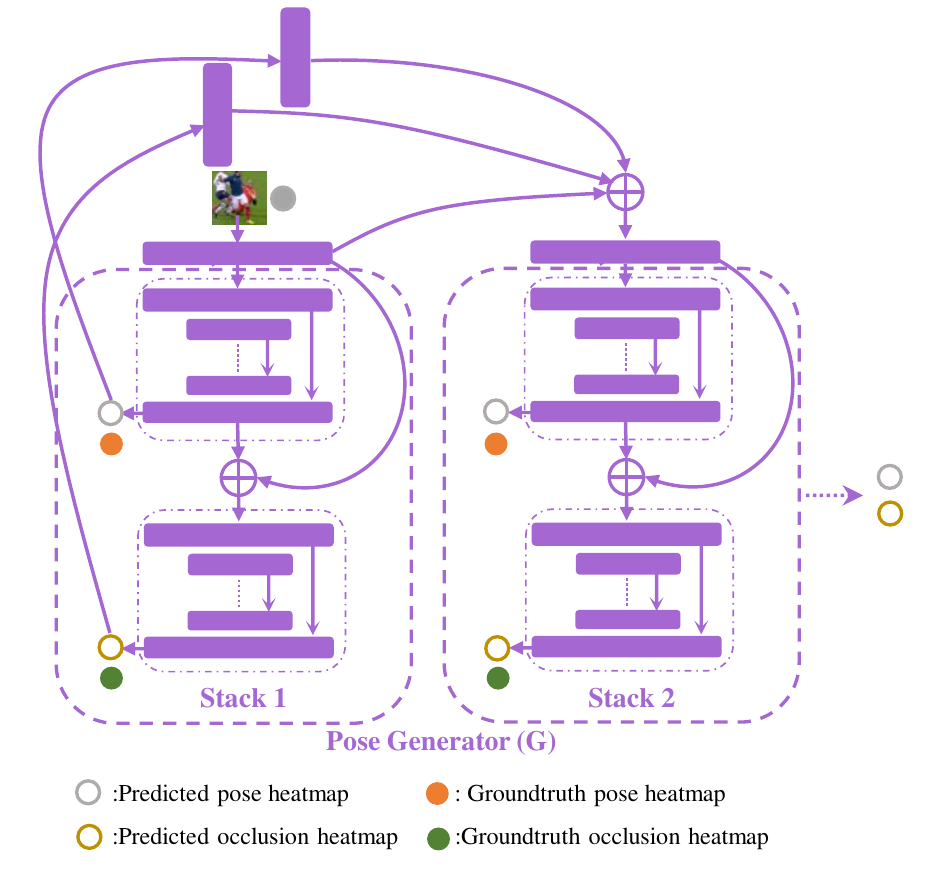}
\caption{Architecture of the multi-task generative network \textit{G}.}
\label{fig:Architecture-of-G}
\end{figure*}

In this section, we introduce the generative network \emph{G} in our framework. Fig.~\ref{fig:Architecture-of-G} presents the architecture of \emph{G}.
Knowledge of whether a body part being occluded clearly offers important information for inferring the geometric information of a human pose. Here, in order to effectively incorporate both pose estimation and occlusion predictions, we propose to tackle the problem with a multi-task generative network.

\def\x{ { \bm x } }

The goal of the multi-task generative network is to learn a function $\mathcal{G}$ which attempts to project an image $\bm{x}$ to both the corresponding pose heatmaps $\bm{y}$ and occlusion heatmaps $\bm{z}$, \emph{i.e.}, $\mathcal{G}(\x) =  \{\hat{\bm{y}}, \hat{\bm{z}}\}$ where $\hat{\bm{y}}$ and $\hat{\bm{z}}$ are the predicted heatmaps.

In addition, as reported in~\cite{conf/cvpr/WeiRKS16}, large contextual regions are important for locating body parts. So the contextual region of a neuron, which is its receptive field, should be large. To achieve this goal, an ``encoder-decoder'' architecture is used.

Besides, for the problem of human pose estimation, local evidence is essential for identifying features for faces and hands. Meanwhile, the final pose estimation requires a coherent understanding of the full body image. To capture this information at each scale, skip connections between mirrored layers in the encoder and decoder are added. Inspired by~\cite{conf/eccv/NewellYD16}, our network is also stacked to provide the network with a mechanism for re-evaluation of initial estimates and features across the entire image. In each module of the \textit{G} net, a residual block~\cite{conf/cvpr/HeZRS16} is used for the convolution operator. Given the original image $\bm{x}$, a basic block of the stacked multi-task generator network can be expressed as follows:
\begin{equation}
\nonumber
\begin{cases}
\{\bm{Y}_{n},\bm{Z}_{n},\bm{X}\}=\mathcal{G}{}_{n}(\bm{Y}_{n-1},\bm{Z}_{n-1},\bm{X})  \quad{\rm if~~}n\geqslant2\\
\{\bm{Y}_{n},\bm{Z}_{n},\bm{X}\}=\mathcal{G}{}_{n}(\bm{X})\quad\quad\quad\quad\quad\quad~~ {\rm if~~} n=1
\end{cases}
\,,
\end{equation}
where $\bm{Y}_{n}$ and $\bm{Z}_{n}$ are the output activation tensors of the $n\textrm{-th}$ stacked generative network for pose estimations and occlusion predictions, respectively. $\bm{X}$ is the image feature tensor, obtained
after pre-processing on the original image through two residual blocks.
Suppose that there are $N$ times stacking of the basic block, then the multi-task generative network can be formulated as:
\begin{equation}
\nonumber
\{\bm{Y}_{N},\bm{Z}_{N},\bm{X}\}=\mathcal{G}_{N}(\mathcal{G}_{N-1}(\cdots(\mathcal{G}_{1}(\bm{X}),\bm{Y}_{1},\bm{Z}_{1}))) \,.
\end{equation}
In each basic block, the final heatmap outputs $\hat{\bm{y}}_{n},\hat{\bm{z}}_{n}$ are obtained from $\bm{Y}_{n}$ and $\bm{Z}_{n}$ by two $1\times 1$ convolution layers with the step size of 1 and without padding. Specifically, the first convolution layer reduces the number of feature maps from the number of feature maps to the number of body parts. The second convolution layer acts as a linear classifier to obtain the final predicted heatmaps.

Therefore, given a training set $\{\bm{x}^{i},\bm{y}^{i},\bm{z}^{i}\}_{i=1}^{M}$ where $M$ is the number of training images, the loss function of our multi-task generative network is presented as:
\begin{equation}
\label{eq:multitaskG}
\mathcal{L}_{G}(\Theta)=\frac{1}{2MN}\sum_{n=1}^{N}\sum_{i=1}^{M}\left(\left\Vert \bm{y}_{n}^{i}-\hat{\bm{y}}_{n}^{i}\right\Vert ^{2}+\left\Vert \bm{z}_{n}^{i}-\hat{\bm{z}}_{n}^{i}\right\Vert ^{2}\right) \,.
\end{equation}
where $\Theta$ denotes the parameter set.

\subsection{Pose Discriminator}\label{subsec:Pose-Discriminator}

To enable the training of the network to exploit priors about the human body joints configurations, we design the pose discriminator \textit{P}. The role of the discriminator \emph{P} is to distinguish the {\it fake} poses (poses do not satisfy the constraints of human body joints) from the {\it real} poses.

It is intuitive that we need local image regions to identify the body parts and the large image patches (or the whole image) to understand the relationships between body parts. However, when some parts are seriously occluded, it can  be very difficult  to locate the body parts. Human can achieve that by  using
prior knowledge and observing both the local image patches around the body parts and relationships among different body parts. Inspired by this, both low-level and high-level information can be important to infer whether the predicted poses are biologically plausible. In contrast to previous work, we use an encoder-decoder architecture to implement the discriminator \emph{P}. Skip connections between parallel layers are used to incorporate both the local and global information.

\begin{figure*}[!t]
\centering
\includegraphics[width=1.34695\columnwidth]{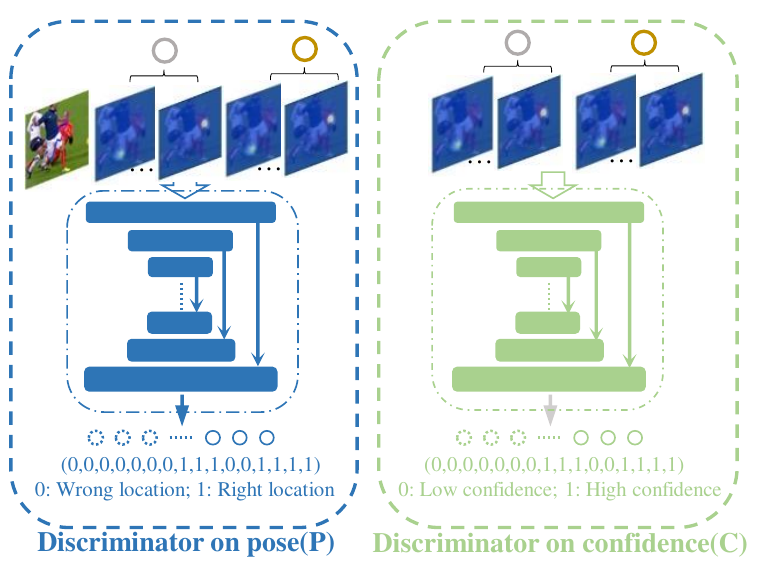}
\caption{Architectures of the discriminator network \textit{P} and \textit{C}.}
\label{fig:Architectures-of-P}
\end{figure*}

Additionally, even when the generative network fails to predict the correct pose locations for a particular image, the predicted pose may be a plausible one, only for a different human body shape. Thus, simply using the pose and occlusion features may still face difficulty in training an accurate \textit{P}.

{\it Such inference should be made by taking the original image into consideration at the same time.} Occlusion information can also be useful in inferring the pose rationality. We use the input RGB image with pose and occlusion maps generated by the \textit{G} net as the input to \textit{P} for predicting whether a pose is reasonable or not for a particular image. The network structure of \textit{P} is shown in Fig.~\ref{fig:Architectures-of-P}.
To achieve this goal, GAN is set in the conditional manner for \textit{P} in our framework. As GANs learn a generative model of data, conditional GANs (cGANs) learn a conditional generative model~\cite{conf/nips/GoodfellowPMXWOCB14}. The objective of a conditional adversarial \textit{P} network is expressed as follows:
\begin{equation}
\label{eq:disP}
\mathcal{L}_{P}(G,P)=\mathbb{E}[\textrm{log}P(\bm{y},\bm{z},\bm{x})]+\mathbb{E}[\textrm{log}(1-(P(G(\bm{x}),\bm{x})-\bm{p}_{\rm  fake}))] \,.
\end{equation}
where $\bm{p}_{\rm  fake}$ is the ground truth pose discriminator label. In traditional GAN,  $\bm{p}_{\rm  fake}$ is simply set as 0. The choice of $\bm{p}_{\rm  fake}$ here will be discussed in detail in Section~\ref{subsec:Geometric-Constrained-Training}.

\subsection{Confidence Discriminator\label{subsec:Confidence-Discriminator}}

By observing the differences between ground truth heatmaps and predicted heatmaps by previous methods, we find that the predicted ones are often
not Gaussian centered because of occlusions and body overlapping. Recalling the mechanism of human vision, even when the body parts are occluded,
we can still confidently locate the body parts. That is mainly because we already acquire  the geometric prior of human body joints. Motivated by this, we design a second  auxiliary discriminator, which is termed Confidence Discriminator (\emph{i.e.}, \textit{C}) to discriminate the high-confidence predictions from the low-confidence predictions. The inputs for \textit{C} are the pose and occlusion heatmaps. The objective of a traditional adversarial \textit{C} network can be expressed as:
\begin{equation}
\label{eq:disC}
\mathcal{L}_{C}(G,C)=\mathbb{E}[\textrm{log}C(\bm{y},\bm{z})]+\mathbb{E}[\textrm{log}(1-(C(G(\bm{x}))-\bm{c}_{\rm  fake}))] \,.
\end{equation}
where $\bm{c}_{\rm  fake}$ is the ground truth confidence label. In traditional GAN,  $\bm{c}_{\rm  fake}$ is simply set as 0. The choice of $\bm{c}_{\rm  fake}$ here will also be discussed in Section~\ref{subsec:Geometric-Constrained-Training}.

\subsection{Training of the Adversarial Networks}\label{subsec:Geometric-Constrained-Training}

In this section, we describe the detailed inputs and outputs of the discriminators, and then explain in detail how \textit{P} and \textit{C} contribute to the accurate pose predictions with structure constraints.

As mentioned in Section~\ref{subsec:Pose-Discriminator} and Section~\ref{subsec:Confidence-Discriminator}, the previous pose estimation networks usually have less confidences in locating the occluded body parts as the local information are neglected. However, if the \textit{G} network can learn to see through the occlusions like the inferences made by human in this situation, it should  achieve high confidences in locating such body parts.

If \textit{G} generates low-confidence heatmaps, \textit{C} will classify the result as ``fake''. As \textit{G} is optimized to deceive \textit{C} that the fakes being real, this process would help \textit{G} to generate high confidence heatmaps even with occlusions presented. The outputs are the confidence scores $\bm{c}$ which in fact corresponds to whether the network is confident in locating body parts.

During training \textit{C}, the real heatmaps are labelled with  a $16\times 1$ (16 is the number of body parts) unit vector $\bm{c}_{\rm real}$. The confidence of the fake (predicted) heatmap should be high when it is close to ground truth and low otherwise. So the fake (predicted) heatmaps are labelled with a   $16\times 1$ vector $\bm{c}_{\rm fake}$ where the elements of $\bm{c}_{\rm fake}$ are the corresponding confidence scores.
\begin{equation}
\nonumber
\bm{c}_{\rm fake}^{i}=\begin{cases}
1 & {\rm if}\thinspace\left\Vert \bm{y}_{i}-\hat{\bm{y}_{i}}\right\Vert <\varepsilon\\
0 & {\rm if}\thinspace\left\Vert \bm{y}_{i}-\hat{\bm{y}_{i}}\right\Vert \geqslant\varepsilon
\end{cases}\,,
\end{equation}
where $\varepsilon$ is the threshold parameter, and $i$ is the $i$-th body part. The range of the output values in \textit{C} is $\left[0,1\right]$.

In the following, we show how to embed the geometric information of human bodies into the proposed \textit{P} network. We observe that, when a part of human body is occluded, the prediction of the un-occluded parts are typically not affected. This may be due to the DCNN's strong ability to learn local features.

In previous works on image translation using GANs, the discriminative network is learned with  real samples being labeled 1 and 0 for fake samples.
For the problem of human pose estimation, we found the network to be difficult to converge by simply setting 0 or 1 as ground truth for a sample. Instead, we designed a novel strategy for pose estimation.

\begin{algorithm}[h!]
\caption{The training process of our method.}
\label{alg:The-training-process}
\small
\begin{algorithmic}[1]{
\REQUIRE {Training images:  $\bm{x}$, the corresponding ground-truth heatmaps \{\textit{$\bm{y}$,$\bm{z}$}\}};
\STATE {Forward \textit{G} by $\{\hat{\bm{y}},\hat{\bm{z}}\}=\emph{G}(\bm{x})$, and optimize \textit{G} according to Eq.~\eqref{eq:multitaskG}};
\STATE {Forward \textit{P} by $\{\hat{\bm{p}}_{\rm fake}\}=\emph{P}(\bm{x},\emph{G}(\bm{x}))$, and optimize \textit{P} net by maximizing the second term in Eq.~\eqref{eq:disP}};
\STATE {Forward \textit{P} by $\{\hat{\bm{p}}_{\rm real}\}=\emph{P}(\bm{x},\bm{y},\bm{z})$, and optimize \textit{P} by maximizing the first term in Eq.~\eqref{eq:disP}};
\STATE {Forward \textit{C} by $\{\hat{\bm{c}}_{\rm fake}\}=\emph{C}(\emph{G}(\bm{x}))$, and optimize \textit{C} by maximizing the second term in Eq.~\eqref{eq:disC}};
\STATE {Forward \textit{C} by $\{\hat{\bm{c}}_{\rm real}\}=\emph{C}(\bm{x},\bm{y},\bm{z})$, and optimize \textit{C} by maximizing the first term in Eq.~\eqref{eq:disC}};
\STATE {Optimize \textit{G} by  Eq.~\eqref{eq:final}};
\STATE {Go back to \textbf{Step 1} until the accuracy of the validation set stop increasing};
\RETURN {\emph{G}}.
}\end{algorithmic}
\end{algorithm}

The ground truth $\bm{p}_{\rm real}$ of a real sample is a $16\times 1$ unit vector. For the fake samples, if a predicted body part is far from the ground truth location, the pose is clearly implausible  for the body configuration in this image. Therefore, the ground truth $\bm{p}_{\rm fake}$ is:
\begin{equation}
\nonumber
\bm{p}_{\rm fake}^{i}=\begin{cases}
1 & {\rm if}\thinspace d_{i}<\delta\\
0 & {\rm if}\thinspace d_{i}\geqslant\delta
\end{cases}\,,
\end{equation}
where $\delta$ is the threshold parameter and $d_{i}$ is the normalized distance between the predicted and ground-truth location of the $i$-$\textrm{th}$ body part. The range of the output values in \textit{P} is also $\left[0,1\right]$. To deceive \textit{P}, \textit{G} will be trained towards the direction to generate heatmaps which satisfy the joints constraints of human bodies. Previous approaches to conditional GANs have found it beneficial to mix the GAN objective with a  traditional loss, such as $\ell_2$ distance~\cite{conf/cvpr/PathakKDDE16}. For our task, it is clear  that we also need to supervise \textit{G} in the training process with the ground truth human poses. Thus, the discriminator still plays the original role, but the generator will not only fool the discriminator but also approximate the ground-truth output in an $\ell_2$ sense as in Eq.~\eqref{eq:disC}. Therefore, the final objective function is presented as follows.
\begin{equation}
\label{eq:final}
\arg \min_G \max_{P,C}  \mathcal{L}_{G}(\Theta)+\mathcal{\alpha L}_{C}(G,C)+\mathcal{\beta L}_{P}(G,P) \,.
\end{equation}
In experiments, in order to make the different components of the final objective function have the same scale, the hyper parameters $\alpha$ and $\beta$ are set to $1/220$ and $1/180$, respectively. Algorithm~\ref{alg:The-training-process} demonstrates the whole training processing as the pseudo codes.

\section{Experiments}\label{sec:Experiments}

\noindent \textbf{Datasets.} We evaluate the proposed method on two widely used benchmarks on pose estimation, \emph{i.e.}, extended Leeds Sports Poses (LSP)~\cite{conf/bmvc/JohnsonE10} and MPII Human Pose~\cite{conf/cvpr/AndrilukaPGS14}. The LSP dataset consists of 11k training images and 1k testing images from sports activities. The MPII dataset consists of around 25k images with 40k annotated samples (about 28k for training, 11k for testing). The figures are annotated with 16 landmarks on the whole body with various challenging directions to the camera. As the annotations of MPII testing set are not provided, we train our model on a subset of training images and evaluate on a held-out validation set about 3000 samples~\cite{conf/cvpr/TompsonGJLB15}. Both datasets provide the visibility of body parts, which is used as the supervision occlusion signal in our method.

\noindent \textbf{Experimental Settings.} According to the detection results, we crop the images with the target human centered at the
images, and warp the image patch to the size of 256$\times$256 pixels. We follow the data augmentation in~\cite{conf/eccv/NewellYD16} by rotation (+/- 30 degrees), and scaling (0.75-1.25) to make the network more robust to different scales and directions. During training for LSP, we use the MPII dataset to augment the training data of LSP, which is a regular routine as done in~\cite{conf/cvpr/WeiRKS16,conf/eccv/InsafutdinovPAA16}.

During testing on the MPII dataset, we follow the standard routine to crop image patches with the given rough position and scale.
 The input resolution to the \textit{G} net is 64$\times$64 pixels. The network starts with a 7$\times$7 convolutional layer with stride 2, followed by a residual modules and a max pooling to drop the resolution down from 256 to 64. Then two residual modules are followed before sending the feature into \textit{G}. Across the entire network all residual modules contain three convolution layers and a skip connection with output of 512 feature maps. The generator is stacked four times if not specially indicated in our experiment. For implementation, we train our model with the Torch7 toolbox~\cite{conf/nips/torch7}. The network is trained using the RMSprop algorithm with initial learning rate of $2.5\times 10 ^{-4}$. The model on the MPII dataset was trained for 230 epochs (about 1 day on a Tesla M40 GPU) and the LSP dataset for 250 epochs (about 1.5 days on a Tesla M40 GPU).

\subsection{Quantitative Results}\label{subsec:Results}

We use the Percentage Correct Keypoints (PCK@0.2) \cite{yang2013articulated} metric for comparison on the LSP dataset which reports the percentage of detection that falls within a normalized distance of the ground-truth for comparisons. For MPII, the distance is normalized by a fraction of the head size~\cite{conf/cvpr/AndrilukaPGS14} (referred to as PCKh@0.5). %

\noindent \textbf{LSP Human Pose.} Table~\ref{tab:Comparisons-of-PCK} shows the PCK performance of our method and previous methods at a normalized distance of 0.2. Our approach outperforms the state-of-the-art across all the body joints, and obtains 2.4\% improvement in average.

\begin{table}[t!]
	\caption{Comparisons of PCK performance with normalized distance being 0.2 on the LSP dataset.}
        \label{tab:Comparisons-of-PCK}
	\small
	\centering
	\setlength{\tabcolsep}{3.0pt}
	\begin{tabular}{|c||c|c|c|c|c|c|c||c|}
		\hline
			{Methods} & \emph{Head} & \emph{Sho.} & \emph{Elb.} & \emph{Wri.} & \emph{Hip} & \emph{Knee} & \emph{Ank.} & \textbf{Mean} \\
		\hline
		\hline
		\cite{Belagiannis2016} & 95.2   & 89.0    & 81.5  &  77.0  &  83.7   &  87.0   & 82.8  &  85.2  \\
		\cite{conf/eccv/LifshitzFU16}  & 96.8  &  89.0  &  82.7  &  79.1  &  90.9  &  86.0   & 82.5   & 86.7  \\
		\cite{conf/iccv/PishchulinAGS13}   & 97.0    & 91.0   & 83.8   & 78.1   & 91.0  &  86.7  &  82.0   & 87.1  \\
		\cite{conf/eccv/InsafutdinovPAA16}   & 97.4   & 92.7   & 87.5  &  84.4  &  91.5  &  89.9   & 87.2  &  90.1  \\
		\cite{conf/cvpr/PishchulinITAAG16}   & 97.8   & 92.5   & 87.0   & 83.9  &  91.5   & 89.9  &  87.2  &  90.1  \\
		\cite{conf/cvpr/WeiRKS16}   & 97.8   & 92.5  &  87.0  &  83.9  &  91.5   & 90.8   & 89.9   & 90.5  \\
		\cite{conf/eccv/BulatT16}   & 97.2   & 92.1   & 88.1   & 85.2   & 92.2   & 91.4  &  88.7   & 90.7  \\
		\hline
		\hline
		Ours & \textbf{98.5}& \textbf{94.0}& \textbf{89.8}& \textbf{87.5}& \textbf{93.9}& \textbf{94.1}& \textbf{93.0}&\textbf{93.1}\\
		\hline
	\end{tabular}
	\vspace{1em}
\end{table}

\begin{figure*}[b!]
\centering
\includegraphics[width=1.5\columnwidth]{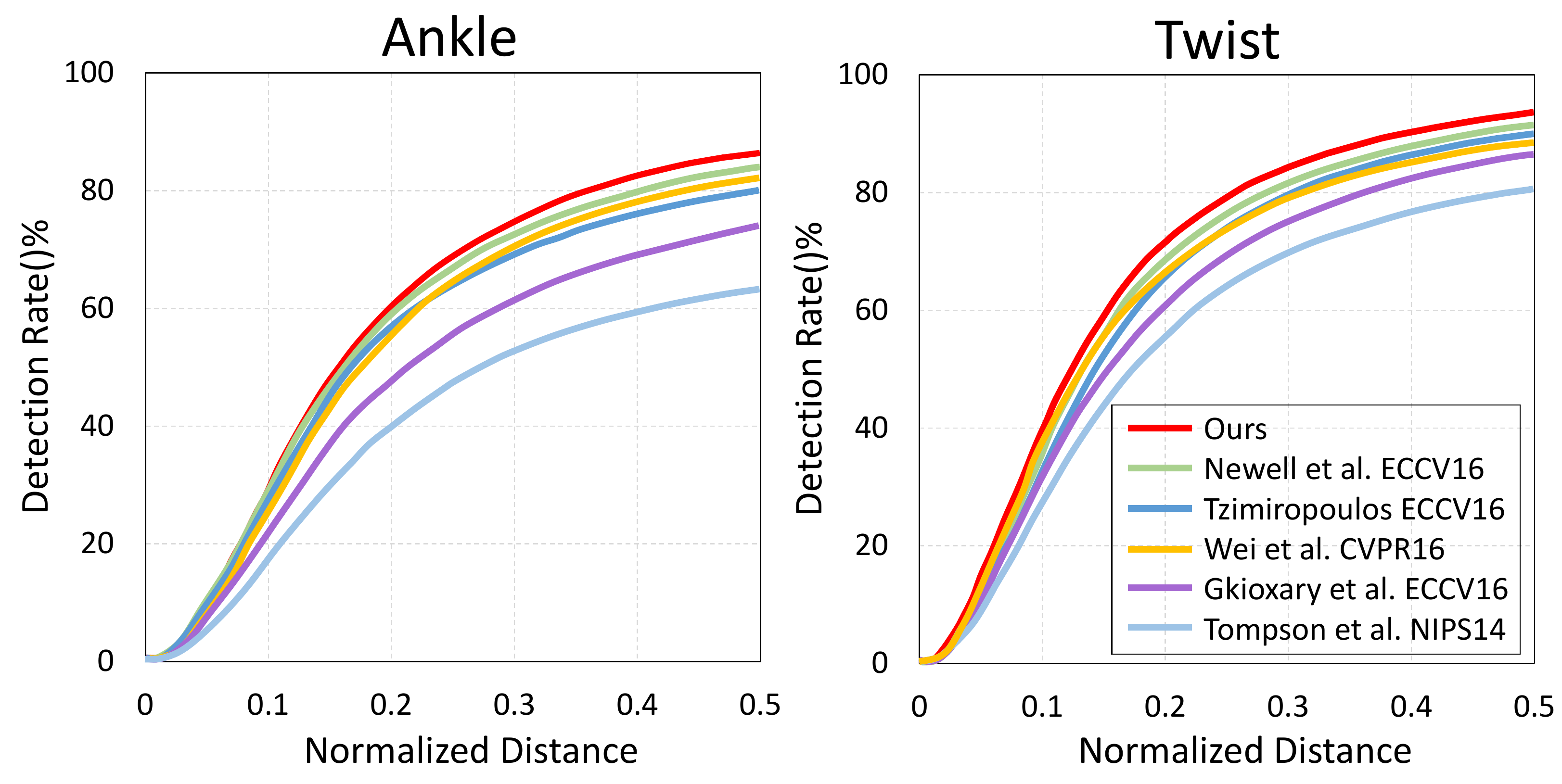}
\caption{PCKh comparison on MPII.}
\label{fig:PCKh-comparison-on}
\end{figure*}

\noindent \textbf{MPII Human Pose.} Table~\ref{tab:Results-on-MPII} and Fig.~\ref{fig:PCKh-comparison-on} reports the PCKh performance of our method and previous methods at a normalized distance of 0.5. ``Ours (-)'' refers to our baseline results by using a four-stacked single-task network without multi-task and discriminators. This network has similar structure but half of stacked layers and parameter numbers compared to~\cite{conf/eccv/NewellYD16}. Our method achieves the best PCKh score of 92.1\% on this data set.

In particular, for the most challenging body parts, \emph{e.g.}, wrist and ankle, our method achieves 1.5\% and 1.9\% improvement compared with the closest competitor respectively, which is significant.

\begin{table}[t!]
 \caption{Results on MPII Human Pose (PCKh@0.5).}
 \label{tab:Results-on-MPII}
 \small
 \centering
 \setlength{\tabcolsep}{3.0pt}
 \begin{tabular}{|c||c|c|c|c|c|c|c||c|}
  \hline
   {Methods} & \emph{Head} & \emph{Sho.} & \emph{Elb.} & \emph{Wri.} & \emph{Hip} & \emph{Knee} & \emph{Ank.} & \textbf{Mean} \\
  \hline
  \hline
  \cite{conf/nips/TompsonJLB14} & 95.8 &  90.3  & 80.5  & 74.3  & 77.6  & 69.7  & 62.8 & 79.6\\
  \cite{conf/cvpr/CarreiraAFM16}  & 95.7 &  91.7 &  81.7 &  72.4 &  82.8  & 73.2  & 66.4 & 81.3\\
  \cite{conf/cvpr/TompsonGJLB15}  & 96.1  & 91.9  & 83.9  & 77.8  & 80.9  & 72.3  & 64.8 & 82.0\\
  \cite{conf/cvpr/HuR16}  & 95.0  & 91.6  & 83.0  & 76.6  & 81.9  & 74.5  & 69.5 & 82.4\\
  \cite{conf/iccv/PishchulinAGS13}  & 94.1  & 90.2  & 83.4  & 77.3  & 82.6  & 75.7  & 68.6 & 82.4\\
  \cite{conf/eccv/LifshitzFU16}  & 97.8  & 93.3  & 85.7  & 80.4  & 85.3  & 76.6  & 70.2 & 85.0\\
  \cite{conf/eccv/GkioxariTJ16}  & 96.2  & 93.1  & 86.7  & 82.1  & 85.2  & 81.4 &  74.1 & 86.1\\
  \cite{BMVC2016}  & 97.2  & 93.9  & 86.4  & 81.3  & 86.8  & 80.6  & 73.4 & 86.3\\
  \cite{conf/eccv/InsafutdinovPAA16}  & 96.8 &  95.2 &  89.3  & 84.4  & 88.4  & 83.4 &  78.0 & 88.5\\
  \cite{conf/cvpr/WeiRKS16}  & 97.8  & 95.0  & 88.7  & 84.0  & 88.4 &  82.8  & 79.4 & 88.5\\
  \cite{conf/eccv/BulatT16}  & 97.9  & 95.1  & 89.9 &  85.3  & 89.4 &  85.7  & 81.7 & 89.7\\
  \cite{conf/eccv/NewellYD16}  & 98.2  & 96.3  & 91.2  & 87.1  & 90.1  & 87.4  & 83.6 & 90.9\\
  \hline
  \hline
    Ours (-)\footnotemark[1]& {98.2}& {96.2}& {90.9}& {86.7}& {89.8}& {87.0}& {83.2}&{90.6}\\
  \hline
  Ours & \textbf{98.6}& \textbf{96.4}& \textbf{92.4}& \textbf{88.6}& \textbf{91.5}& \textbf{88.6}& \textbf{85.7}&\textbf{92.1}\\
  \hline
 \end{tabular}
\footnotemark[1]{\footnotesize{Our baseline using single-task network without multi-task and discriminators}}
\end{table}

\begin{figure*}[!t]
\centering
\includegraphics[width=0.99\textwidth]{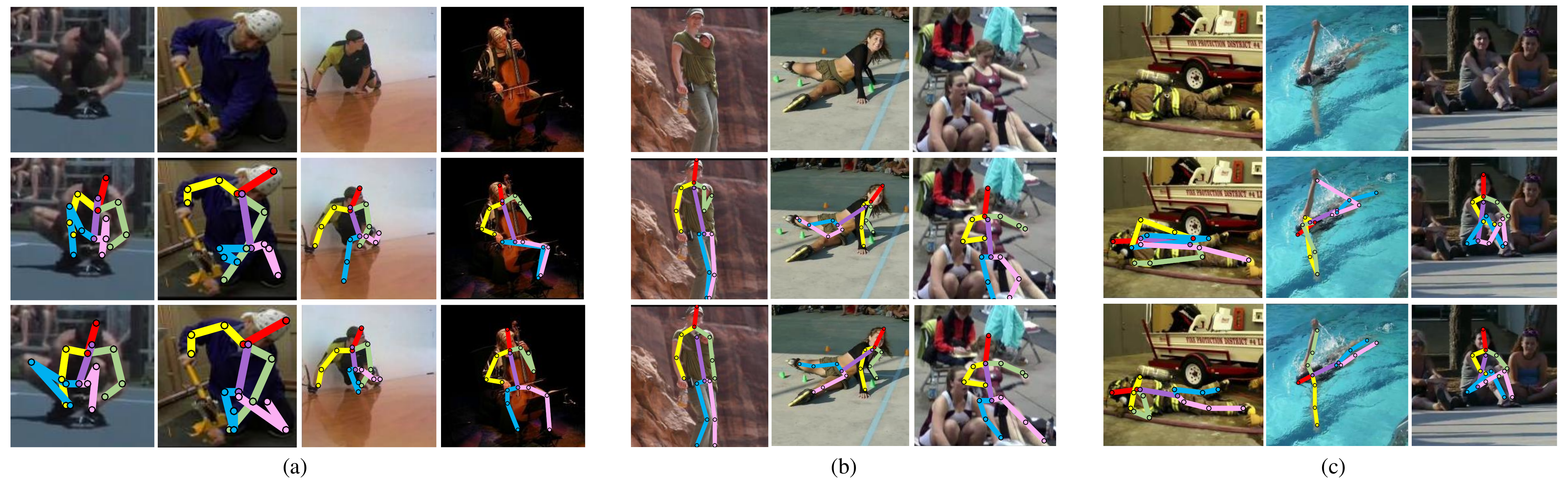}
\caption{Prediction samples on the MPII test set. The first row: original images. The second row: results by stacked hourglass network (HG)~\cite{conf/eccv/NewellYD16}. The third row: results by our method. (a)-(c) stand for three kinds of failure with HG.}
\label{fig:Prediction-samples-on}
\end{figure*}

\subsection{Quanlitative Comparisons}\label{subsec:Comparison-of-Predicted}

To gain insights on how the proposed method accomplish the goal of setting the pose estimations  within the geometric constraints,
we visualize the predicted poses on the MPII test set compared with a 2-stacked hourglass network (HG)~\cite{conf/eccv/NewellYD16}, as demonstrated in Fig.~\ref{fig:Prediction-samples-on}. For fare comparison, we also use a 2-stacked network in this section.
We can see that our method gains a better understanding of the human body image which leads to less strange locations.

In (a), the human body is highly twisted or partly occluded, which results in some invisible body limbs in the image. In these cases, HG fails to understand some poses while our method succeeds. This may be because of the ability of occlusion prediction and shape prior
learned  the in the training process. In (b), HG locates some body parts to the nearby positions with the most salient features. This indicates that HG has learned excellent features about body parts. However, without human body structure awareness, this may locate some body parts to the surrounding area instead of the right one. In (c), due to lack of body configuration constraints, HG produces poses with deviated  twisting across body limbs. As we have implicitly embedded the body constraints into our discriminator, our network succeeds in predicting the correct body location even under some difficult situations.

\begin{figure*}[h!]
\centering
\includegraphics[width=1.99\columnwidth]{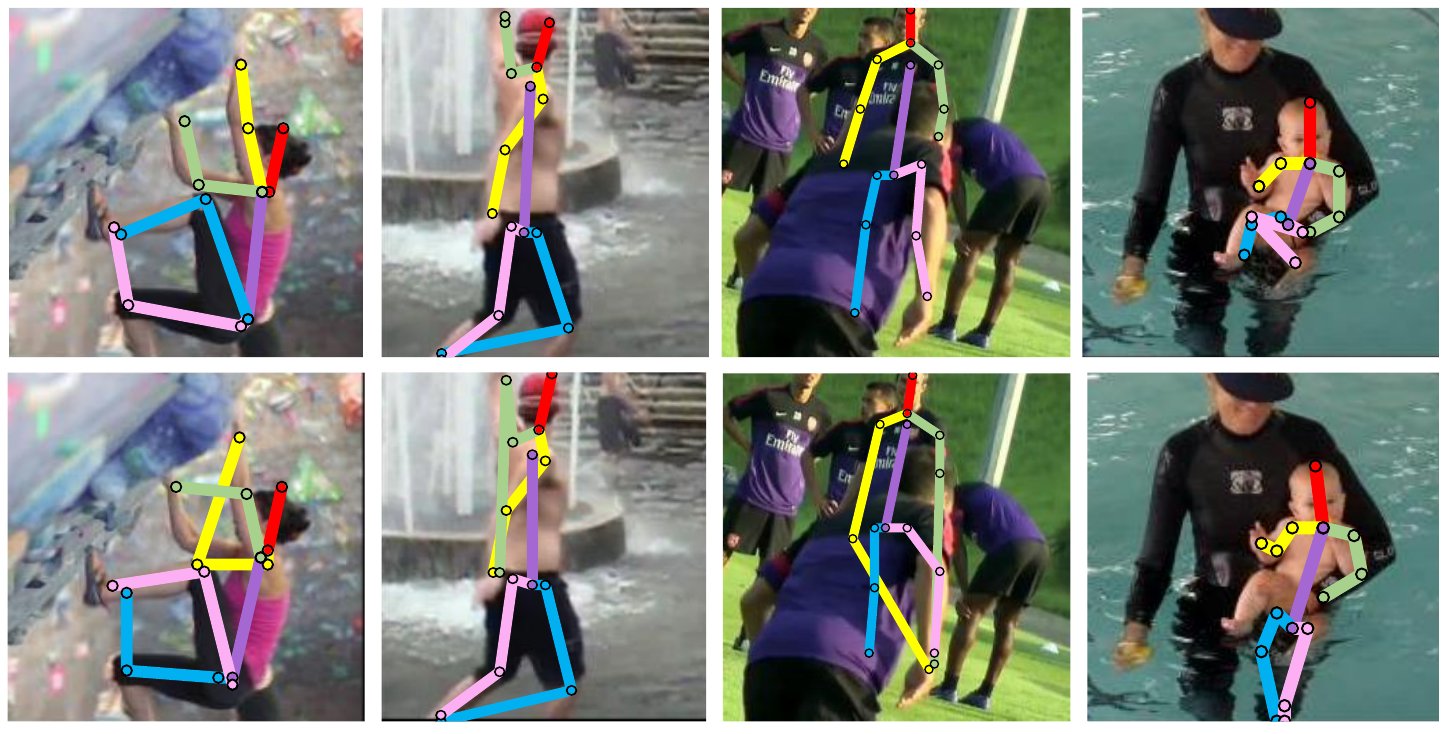}
\caption{Failure cases caused by body part at the edge (the first-second columns), overlapping people (the third column) and invisible limbs (the fourth column). The results on the first and second rows are generated by our method and HG~\cite{conf/eccv/NewellYD16}, respectively.}
\label{fig:Failure-cases-caused}
\end{figure*}

On the other hand, we also show some failure examples of our method on the MPII test set in Fig.~\ref{fig:Failure-cases-caused}. As shown in Fig.~\ref{fig:Failure-cases-caused}, our method may fail in  some challenging cases with twisted limbs at the edge, overlapping people and occluded body parts. In some cases, human may also fail to figure out the correct pose at a glance. However, even when our method fails in this situations, our method also achieves more reasonable poses compared to previous method. In these complicated cases, previous method may generate some poses which are not within the manifold of human pose structure as shown in the first row of Fig.~\ref{fig:Failure-cases-caused}. When the previous network fails to find high-confidence locations around the centered person, it  shifts to the surrounding area where the local features matches the trained feature best. Lacking of shape constrain finally results in some absurd poses.

\subsection{Occlusion Analysis}\label{subsec:Comparison-of-Predicted-1}

Here we present  a detailed analysis of the outputs of the networks when input images are occluded.

\begin{figure*}[h!]
\centering
\includegraphics[width=1.99\columnwidth]{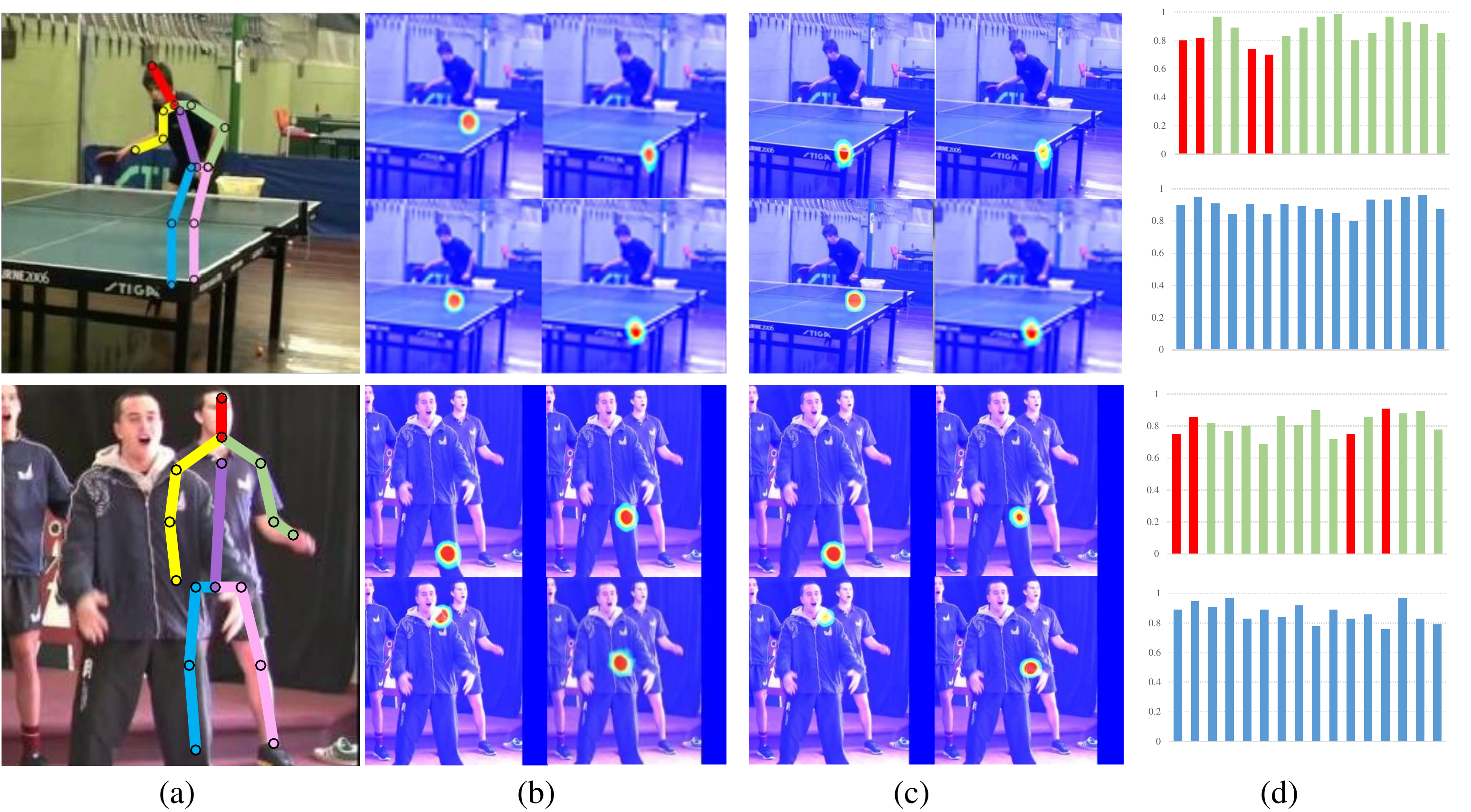}
\caption{(a) Input images with predicted poses; (b) Predicted pose heatmaps of four occluded body parts; (c) Predicted occlusion heatmaps of four occluded body parts; (d) Outputs values of \textit{P} (in blue) and \textit{C}(in green). Red columns in the output of \textit{C} correspond to values of the four occluded body parts.}
\label{fig:Examples-inputs-and}
\end{figure*}

First, two examples with some body parts occluded are given in Fig.~\ref{fig:Examples-inputs-and}. In the first sample, two legs of the person are totally occluded by the table. In the corresponding occlusion maps, the occluded part are well predicted. Despite of the occlusions, the pose heatmaps generated
by our method are mostly clear and Gaussian centered. This results in high scores in both pose prediction and confidence evaluation despite of occlusions.

In the second image, half part of the person is overlapped by the person ahead of him. Our method also succeeds to yield the correct pose locations with clear heatmaps. Occlusion information is also well predicted for the occluded parts. As shown in the columns in red, although the confidence scores of the occluded body parts are comparatively low, they remain an overall high level. This shows that as our network has learned  some human body priors  during training. Thus it has the ability to predict reasonable poses even under some occlusions. This verifies our motivation of designing the discriminators with GANs.

Next, we compare the performance of our method under occlusions with a stacked hourglass network~\cite{conf/eccv/NewellYD16} as the strong baseline. In the validation set of MPII, about 25\% of the elbows and wrists with annotations are labeled invisible. We show the results of elbows and wrists with visible samples and invisible samples in Table~\ref{tab:Detection-rates-of}. For body parts without occlusions, our method improves the baseline by about 0.8\% of detection rate. However, {\it our method improves the baseline by 3.5\% and 3.6\% of detection rates on the invisible twists and elbows. This shows the advantage of our method in dealing with body parts with occlusions.}

\subsection{Ablation Study}

\begin{figure*}[h!]
\centering
\includegraphics[width=1.285\columnwidth]{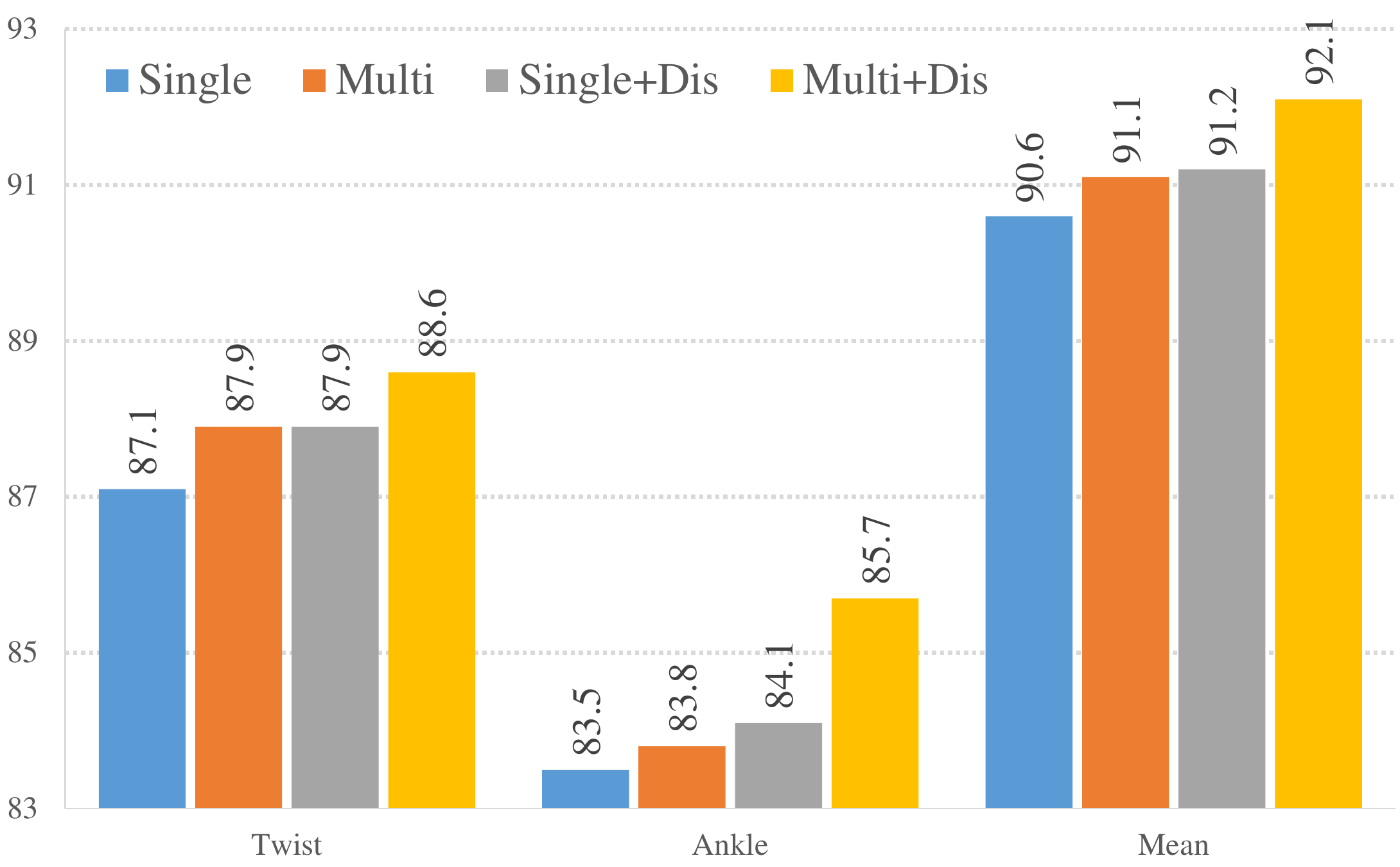}
\caption{Ablation study: PCKh scores at the threshold of 0.5.}
\label{fig:9}
\end{figure*}

To investigate the efficacy of the proposed multi-task generator network and the  discriminators designed for learning human body priors, we conduct ablation experiments on the validation set of the MPII Human Pose dataset. A four-stacked single-task generator without occlusion is used as the baseline. The overall result is shown in Fig.~\ref{fig:9}. We give analysis to two components in our method: the multi-task manner and discriminators.

\begin{table}[b!]
 \caption{Detection rates of visible and invisible elbows and wrists.} \label{tab:Detection-rates-of}
 \small
 \centering
 \renewcommand\arraystretch{1}
 \begin{tabular}{|c||cc|cc|}
  \hline
  \multirow{2}{*}{Methods} & \multicolumn{2}{c|}{Visible} & \multicolumn{2}{c|}{Invisible}\\
\cline{2-5} & \emph{Twist} & \emph{Elbow} & \emph{Twist} & \emph{Elbow} \\
  \hline
  \hline
 ~\cite{conf/eccv/NewellYD16} & 93.6  &  95.1  &  67.2  &  74.0 \\
  \hline
  Ours &  \textbf{94.5}  &  \textbf{95.9}   & \textbf{70.7}  &  \textbf{77.6}  \\
  \hline
 \end{tabular}
\end{table}

\textbf{Multi-task.} We compare the four-stacked multi-task generator with the baseline. The networks are trained simply by removing the generators (\emph{i.e.}, no GANs). By using the occlusion information, the performance on the MPII validation set increases 0.5\% compared to the baseline model. This shows that the multi-task structure helps the network to understand the poses.

\textbf{Discriminator with Single-task.} We also compare the four-stacked single-task generator trained with discriminators with the baseline. The networks are trained by removing the part for the occlusion heatmaps. Discriminators also receive inputs without occlusion heatmaps. By using the body-structure-aware GANs, the performance on the MPII validation set increases by 0.6\% compared to the baseline model. This shows that the discriminators  contribute in pushing the generator to produce more reliable pose predictions.

In general, individually adding the multi-task or discriminator both increase the accuracy of location. But using them separately results in 0.6\% and 0.5\% improvement respectively, while using both produces an improvement of 1.5\%. The may be due to the reliability of \textit{P} and \textit{C }  on sufficient feature to discriminate the results. Occlusion features obviously can help to understand the image and the generated pose for the discriminators.

\section{Conclusions}

In this paper, we proposed a novel conditional adversarial network for pose estimation, termed Adversarial PoseNet, which trains a multi-task pose generator with two discriminator networks. The two discriminators function like an expert who distinguishes reasonable poses from unreasonable ones. By training the multi-task pose generator to deceive the expert that the generated pose is real, our network is more robust to occlusions, overlapping and twisting of human bodies. In contrast to previous work using DCNNs in human pose estimation, our network is able to alleviate  the risk of locating the human body part onto the  matched features  without consideration of human body priors. 

Although we need to train three sub-networks (\textit{G}, \textit{P}, \textit{C}), we only need to use \textit{G} net during testing. With a small computation overhead, we achieve considerably better results on two popular benchmark datasets. 
We have also verified that our network can produce human poses which are mostly within the manifold of human body shape.

The method developed here can be immediately applied to other shape estimation problems such as face landmark detection using DCNNs. More significantly, we believe that the use of GANs as a tool to predict structured  output  or enforcing output dependency can be further developed to much more general structured output learning.

{\small
\bibliographystyle{ieee}
\bibliography{draft_2}
 }

\end{document}